# Point Cloud Colorization Based on Densely Annotated 3D Shape Dataset

Xu Cao and Katashi Nagao

Department of Intelligent Systems, Graduate School of Informatics
Nagoya University, Nagoya, Japan
sou@nagao.nuie.nagoya-u.ac.jp, nagao@i.nagoya-u.ac.jp

**Abstract**—This paper introduces DensePoint, a densely sampled and annotated point cloud dataset containing over 10,000 single objects across 16 categories, by merging different kind of information from two existing datasets. Each point cloud in DensePoint contains 40,000 points, and each point is associated with two sorts of information: RGB value and part annotation. In addition, we propose a method for point cloud colorization by utilizing Generative Adversarial Networks (GANs). The network makes it possible to generate colours for point clouds of single objects by only giving the point cloud itself. Experiments on DensePoint show that there exist clear boundaries in point clouds between different parts of an object, suggesting that the proposed network is able to generate reasonably good colours. Our dataset is publicly available on the project page[1].

## 1 Introduction

Today, there are multiple devices and applications that have introduced 3D objects and scenes in different areas, such as architecture, engineering, and construction. 3D digitization of the real world is becoming essential for developing a variety of applications such as autonomous driving, robotics, and augmented/virtual reality. Medical and cultural fields, and many others, have benefitted from 3D digitization; examples include prosthesis construction adapted to the anthropometry of each patient or making virtual tours through historic buildings.

A point cloud, which is a 3D representation of real-world objects, consists of a set of points with XYZ-coordinates. A point cloud can be obtained by range-sensing devices such as LiDAR (light detection and ranging). LiDAR has a 360-degree field of view but can only provide sparse depth information. In the case of indoor scene capture, a LiDAR-based 3D scanner solved this problem by vertically rotating LiDAR to acquire sparse point clouds from different orientations and merging them into a dense point cloud. However, the point clouds obtained by LiDAR do not have colour information, making it hard to utilize them in some applications. This does not necessarily mean we need to complete point clouds with accurate colour information. In Nagao et al.'s study

---

1. http://rwdc.nagao.nuie.nagoya-u.ac.jp/DensePoint

[12], an indoor scene is represented as a coloured point cloud and imported into a virtual reality application such as a simulation of a disaster experience. In the case of virtual reality, there is no need for the colours of objects to be exactly the same as those of the real world.

We also require object part information, such as head and body information, because it is impossible to properly transform objects (e.g., disassemble them due to the impact of, for example, an earthquake) in simulation without object part information.

To handle the problems of object colorization and part segmentation, first of all, we constructed DensePoint, which is a dataset that contains the shape, colour, and part information of the object by using a 3D point cloud. DensePoint is an extension of the information in the ShapeNet [2] and ShapeNetPart [25] published datasets.

In this paper, we tackle an automatic point cloud colorization problem as the first application of the DensePoint dataset. That is, given a point cloud without colour information, our goal is to generate a reasonably good colorized point cloud. We take inspiration from pix2pix [8], in which images from one domain are translated into another domain, resulting in interesting applications such as monochrome image colorization. To the best of our knowledge, the point cloud colorization task has not been challenged yet. We think the reasons are the lack of a coloured point cloud dataset and the intractable properties of point clouds. As mentioned earlier, we first constructed a richly annotated point cloud dataset and then adopted recent advances of Generative Adversarial Networks (GANs) to handle the point cloud colorization problem.

## 2  Related Work

### 2.1  3D Shape Repository

A key factor of the success of data-driven algorithms is large-scale and well-annotated datasets. Early efforts in constructing 3D model datasets either do not pay attention to the numbers of models [3] or do not focus on annotating the model [17]. Wu et al.'s study [22] demonstrated the benefit of a large 3D dataset in training convolutional neural networks for 3D object classification tasks, and the dataset named ModelNet has been one benchmark for 3D object classification. The emergence of the large-scale 3D shape repository ShapeNet [2] has facilitated researches in computer graphics, computer vision, and many other fields. ShapeNet provides over 55k single clean mesh models of multiple categories collected from public online sources and other datasets and organizes these models under WordNet taxonomy. Several studies contribute augmentations to the original ShapeNet. ShapeNetPart [25] adds part annotations to 3D shapes of ShapeNet while ObjectNet3D [23] aligns objects in images with 3D shape instances and their pose estimations. In Shao et al.'s study [16], the physical attributes of real-world objects, such as weights and dimensions, are collected from the Internet and then assigned to 3D shapes.

## 2.2 Deep Learning for Point Clouds

Because of the unstructured data format, it is hard for point cloud classification to benefit from the advances of convolutional operation, which has become a standard approach in image classification, segmentation, or object detection tasks. PointNet [14] was the first neural network to address point cloud classification and segmentation by applying point-wise convolution and using a symmetric function to aggregate feature-wise information. PointNet++ [15] improved on PointNet by capturing local structure in a hierarchical way. Other point cloud classification attempts focus on modifying convolutional operation to adapt it to the special format of point clouds [9, 19].

## 2.3 Generative Adversarial Nets for 3D Shape Synthesis

With recent advances of Generative Adversarial Networks [6, 7], many studies contribute to 3D shape generation and completion in a data-driven approach. 3D-GAN [21] and 3D-IWGAN [18] generate volumetric objects by learning a probabilistic mapping from latent space to volumetric object space. 3D shape reconstruction is another task in which a complete 3D object is reconstructed from a partial observation or data in a different modality, such as partial depth view [24], image [18, 21], or multi-view sketches [11]. Even a complete indoor scene can be reconstructed from partial observations, such as an incomplete 3D scan [4] or a single depth view [19]. While these studies focus on volumetric representations of 3D objects, recent studies have also addressed the problem of generating a 3D object in the form of point clouds [5, 10, 13].

# 3 Point Cloud Dataset Construction

In this section, we describe our procedure for constructing a dataset containing densely sampled point clouds with each point associated with RGB colour and a part label.

## 3.1 Data Source

We use ShapeNet [2] and ShapeNetPart [25] as our data source, of which the former provides over 50,000 mesh models across 55 categories and the latter comprise over 30,000 per-point labelled point clouds from 16 categories. As ShapeNetPart is an extension of ShapeNet, both datasets contain the same 3D objects yet in a different modality. We focus on the intersection of the two datasets, a set of over 10,000 3D models, and combine the information of 3D models in different modalities.

## 3.2 Point Cloud Sampling and Alignment

We first uniformly sample points from the surface of mesh objects that have texture in ShapeNet. For each mesh, we densely sample 40,000 points (Fig. 1).

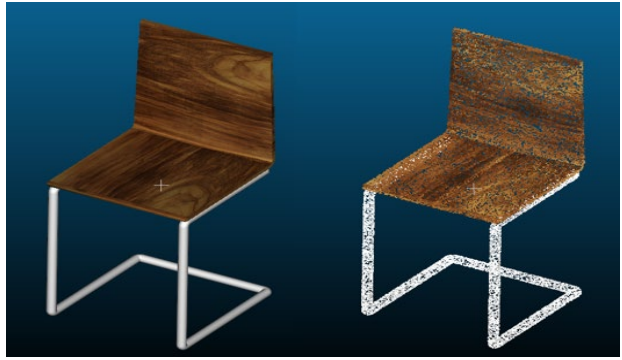

**Fig. 1.** Sampled point cloud visualization. Left: mesh object of chair from ShapeNet. Right: corresponding sampled point cloud.

The alignment process (Fig. 2) consists of 4 separate steps. First, the coloured point clouds are rotated such that the orientations of the point cloud pairs are the same. Second, the centres of the bounding boxes are matched so that the offset of the point cloud pairs disappears. Third, the scales of the point cloud pairs are adjusted to make sure they are the same size. Finally, for point clouds pairs that don't align well, we manually adjust them.

To evaluate the degree of alignment between the point cloud pairs, we utilize the one-sided Hausdorff distance. The one-sided Hausdorff distance between a set of points A and another set of points B is the smallest distance such that for every point of A, there must exist at least one point of B within the distance. Formally, the distance is defined as:

$$d(A, B) = max_{a \in A}\{min_{b \in B}\{\| a - b \|_2\}\}$$

where a and b represent a single point of A and B, respectively. In our case, a and b are vectors of 3 elements representing x, y, and z coordinates in Euclidean space.

After each step, we compute the one-sided Hausdorff distance for all point cloud pairs and then compute the average distance for each category (Fig. 3). We found that for all 16 categories, the average one-sided Hausdorff distance decreases as the point cloud pairs are progressively processed, which verifies the effectiveness of the process.

Finally, we use the one-sided Hausdorff distance to check whether abnormal operation happened in previous steps by computing the one-sided Hausdorff distance between the point cloud pairs after each step. Ideally, the distance should keep decreasing as the alignment process is going on since each step makes the point cloud pairs more similar. We consider point cloud pairs where the distance does not decrease during the process as abnormal point cloud pairs and manually check and adjust the point cloud pairs.

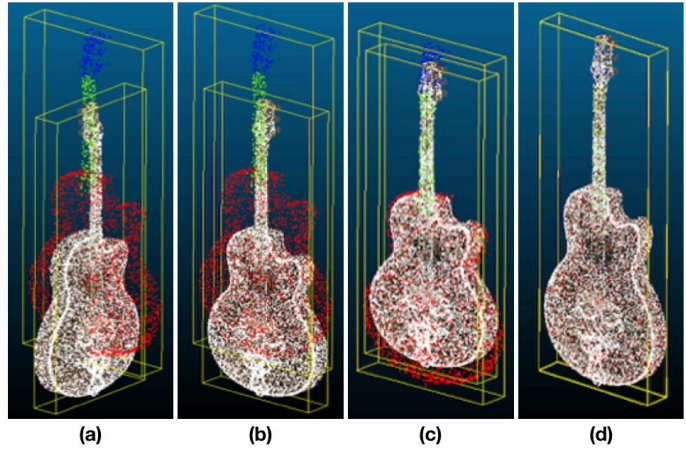

**Fig. 2.** Illustration of alignment process of a point cloud pair. Denser one is sampled from mesh model of ShapeNet, and its colour represents RGB value, while sparser one is from ShapeNetPart, and different colour of points means different parts of object. (a) Original point cloud pair. Note that neither orientation nor scale is same although they originate from same mesh object. (b) After rotation, orientation of point cloud pair became same. (c) Centres of bounding boxes is matched. (d) Scale of point cloud pair is adjusted to be same.

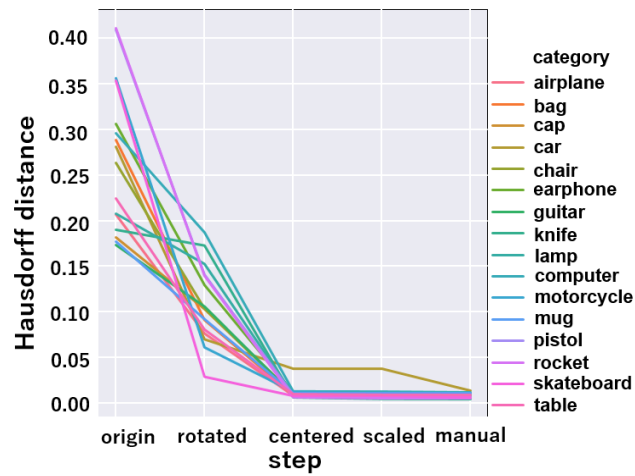

**Fig. 3.** Change of average one-sided Hausdorff distance for all 16 categories. X-axis represents different processing steps, and Y-axis represents one-sided Hausdorff distance. Point cloud pairs from all categories achieve low Hausdorff distance at the end of our proposed procedure.

### 3.3 Label Annotation Transfer

After each point cloud pair is aligned, the problem becomes how to transfer label annotation from the sparse point cloud to the dense point cloud. A prior observation is that points with the same part label are spatially close and clustered, which means there is a high probability for a point to have the same label as those around it. Therefore, we adopt the K-nearest neighbours algorithm for point-wise classification, in which the training data is the points from the point cloud with label annotation, and the test data is the points without label annotation. We train each classifier for every point cloud pair, resulting in over 10,000 classifiers. To find out the best classifier, we consider the combination of two hyperparameters. The first one is K, which is the number of nearest points in the training data to be searched for. The second one is the weight strategy associated with the K-nearest points when voting for the test point label. We search for K from 1 to 17 with a step of 2 and chose two different weight strategies, whether the weights are all the same as the weight of 1 or are inverse to the distance from the query point to the nearest point.

This search strategy results in 18 hyperparameter settings. To decide the best classifier among the 18 settings, we adopt 10-fold cross-validation, which is a standard technique to evaluate trained classifiers. In Table 1, we report the average best validation accuracy of point clouds in each category, from which we can see that all classifiers achieve over 95 percent prediction accuracy. After the best classifiers are decided, we deploy them on test point clouds.

**Table 1.** KNN average best validation accuracy for each category

| Category | Guitar | Knife | Pistol | Lamp | Chair | Table | Mug | Car |
|---|---|---|---|---|---|---|---|---|
| Accuracy | 98.8 | 98.9 | 98.7 | 99.2 | 97.6 | 98.7 | 99.5 | 95.5 |
| Category | Bag | Cap | Earphone | Laptop | Skateboard | Rocket | Motorbike | Plane |
| Accuracy | 99.4 | 98.8 | 98.7 | 98.7 | 98.7 | 97.4 | 96.0 | 96.1 |

### 3.4 Dataset Statistics

The detailed statistics of the dataset are summarized in Table 2. We demonstrate examples of each category from our dataset in Fig. 4.

**Table 2.** DensePoint Dataset Statistics

| Category | Guitar | Knife | Pistol | Lamp | Chair | Table | Mug | Car |
|---|---|---|---|---|---|---|---|---|
| No. of instances | 611 | 266 | 166 | 790 | 1998 | 3860 | 66 | 402 |
| No. of Part labels | 3 | 2 | 3 | 4 | 4 | 3 | 2 | 4 |
| Category | Bag | Cap | Earphone | Laptop | Skateboard | Rocket | Motorbike | Plane |
| No. of instances | 57 | 31 | 36 | 338 | 127 | 29 | 159 | 1492 |
| No. of Part labels | 2 | 2 | 3 | 2 | 3 | 3 | 6 | 4 |

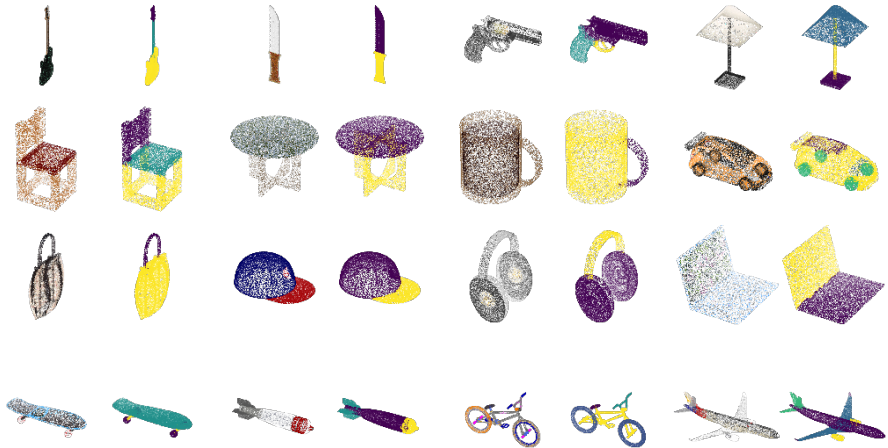

**Fig. 4.** One example of each category from our DensePoint dataset. Each point cloud contains 40,000 points. Left image of each pair is represented by RGB value, and right image is same point cloud represented by part label.

## 4   Point Cloud Colorization

In this section, we explain the architecture of the network, the experiment and the result of point cloud colorization.

### 4.1 Network architecture

We utilize the adversarial scheme of pix2pix [8] and repurpose PointNet [14] segmentation network to colour regression. The architecture of our proposed network is illustrated in Fig. 5. It comprises two neural networks, named generator and discriminator. For the generator architecture, we modify the segmentation version of PointNet, which applies a convolutional operation point by point and then summarizes global information into a vector feature by feature. To accomplish point-wise classification, the global information vector is copied and concatenated with each point-wise feature vector from previous intermediate layer outputs. The activation function of the final layer is a Tanh non-linearity, thus alternating its function from point cloud segmentation to colour regression. For the discriminator architecture, we modify the classification version of PointNet by setting the number of neurons of the output layer to 1, which outputs the probability of the input coloured point cloud being real.

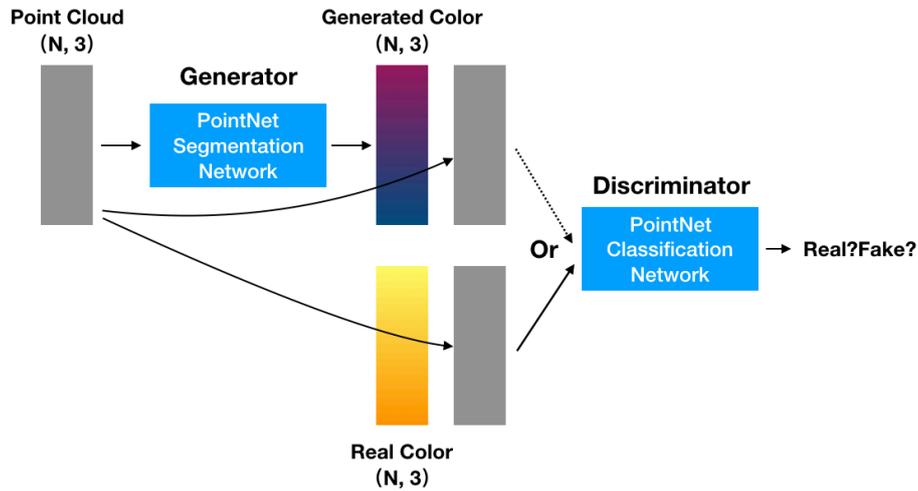

**Fig. 5.** Our generative adversarial network's architecture. Generator, modified from PointNet segmentation network, predicts point-wise colour for N x 3 input point clouds. The predicted colour concatenated with the point cloud, along with the ground truth coloured point cloud, is fed into the discriminator. Discriminator, modified from PointNet classification network, distinguish between

### 4.2 Objective Function

The goal of the generator is to generate realistic point-wise colours for point clouds that are difficult for the discriminator to distinguish from the real coloured point clouds, while the goal of the discriminator is to enhance its own ability to distinguish real colours from generated or fake colours. The optimal situation would be a Nash equilibrium in which neither the generator could fool the discriminator by providing realistic samples nor could the discriminator distinguish real samples from fake samples.

Following pix2pix [8], we utilize a combination of conditional GAN loss and L1 loss, in which conditional GAN loss is defined as:

$$L_{cGAN}(G, D) = \mathbb{E}_{x,y}[log(x, y)] + \mathbb{E}_{x,z}\left[log\left(1 - D(x, G(x, z))\right)\right],$$

and L1 loss is defined as:

$$L_{l_1}(G) = \mathbb{E}_{x,y,z}[\| y - G(x, z) \|_1],$$

where in our case x is the input N x 3 tensor representing a point cloud, and y is the output N x 3 tensor of generator representing point-wise RGB colour. Note that in traditional GANs, z is a random vector input to the generator, which ensures the variation of the output. In our case, we keep the dropout layer at test time so that there is variation of the generated colour for the point clouds.
The final object function is:

$$G^* = \arg min_G \, max_D L_{cGAN}(G, D) + \lambda L_{l_1}(G),$$

where the generator G tries to minimize the combination of conditional GAN loss and L1 loss, and the discriminator D just tries to maximize the object function. λ is a hyperparameter to adjust the importance of the L1 loss relative to the conditional GAN loss.

### 4.3 Experiment and Results

We split all the data into training/test sets following ShapeNet's setting. We train each network on a training dataset for every category and test the network on the corresponding test set. λ is set to 10, and we use an Adam solver for optimizing both the generator and discriminator with a learning rate of 0.0001 for the discriminator and 0.001 for the generator. The optimization steps between the discriminator and the generator are alternate. The imbalance of the generator and the discriminator usually leads to a vanishing gradient and training failure, we adopt a simple strategy to alleviate the problem. Whenever the probability of the discriminator judging the real coloured point cloud to be real is higher than 0.7, we skip training the discriminator this round and jump forward to train the generator until the probability is lower than 0.7. The batch size is 8, and we train our networks for 200 epochs.

We demonstrate our test results in Fig. 6. We found that our proposed network is able to generate reasonably good and beautiful colours for point clouds. Another surprising finding is that the network tends to learn to colorize different parts with different colour patterns by itself even though we did not explicitly provide any information related to the object parts. We observe this phenomenon in almost every category, suggesting that it is not just sampling error and is worth studying further.

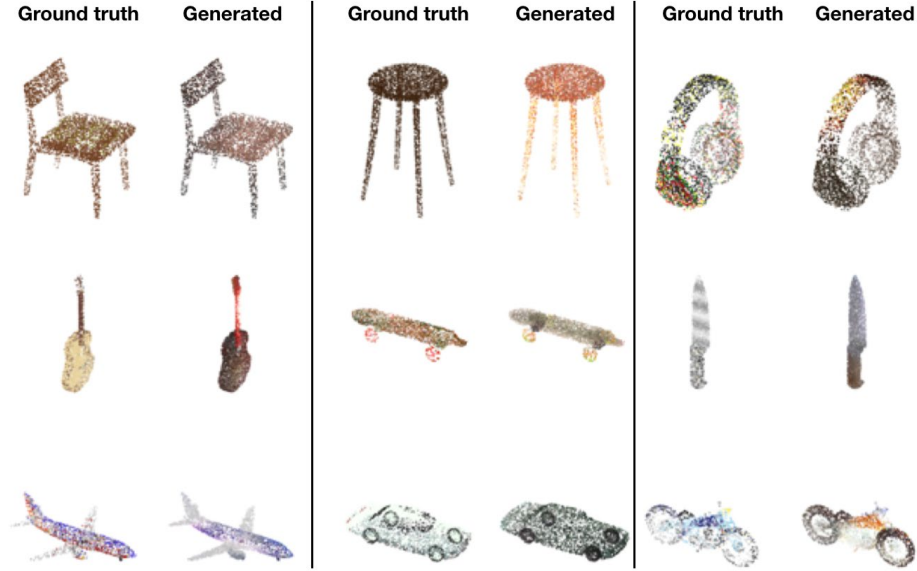

**Fig. 6.** Colorization results on test dataset. Left image in pair is ground-truth coloured point cloud while right image in pair is colorized point cloud. Note that during whole training and test process, we did not give network any information about object parts, but there exist clear boundaries between different parts of a single object.

## 5    Conclusion and Future Work

In this study, we introduce DensePoint, a point cloud dataset comprising over 10,000 single objects across 16 categories, with each point associated with an RGB value and a part label. We also proposed a GAN-based neural network for point cloud colorization task, in which only the point cloud is fed into the network. Clear boundaries between different parts in colourized point clouds indicate that our network is able to generate reasonably good colours for a single object point cloud even if we do not give the network part label information.

Future work includes refining the quality of the label annotations of points as we observe the fact that around the boundary of two sets of points from different parts, there exist some vague and wrong annotations. Another area of future work is exploring the tasks that could be accomplished by utilizing this dataset, such as predicting point-wise part labels while generating the colour for the point clouds at the same time.